\pdfoutput=1
\documentclass[10pt, a4paper]{article}
\usepackage{lrec}

\usepackage{titlesec}
\titleformat{\section}{\normalfont\large\bf\center}{\thesection.}{1em}{}
\titleformat{\subsection}{\normalfont\SmallTitleFont\bf\raggedright}{\thesubsection.}{1em}{}
\titleformat{\subsubsection}{\normalfont\normalsize\bf\raggedright}{\thesubsubsection.}{1em}{}
\renewcommand\thesection{\arabic{section}}
\renewcommand\thesubsection{\thesection.\arabic{subsection}}
\renewcommand\thesubsubsection{\thesubsection.\arabic{subsubsection}}

\usepackage{graphicx}
\usepackage{tabularx}
\usepackage{soul}
\usepackage{relsize}
\usepackage{booktabs}
\usepackage{multirow}
\usepackage{xspace} 
\usepackage{amsmath}

\usepackage[T1]{fontenc}
\usepackage[utf8]{inputenc}

\usepackage{tikz}
\usepackage{tikz-dependency}
\usetikzlibrary{fit,patterns,shapes,arrows,calc,shadows,positioning}

\usepackage{covington}

\usepackage{hyperref}
\usepackage{xstring}

\usepackage{todonotes}

\newcommand{\norex}[1]{\textit{#1}}
\newcommand{\eng}[1]{`#1'}

\newcommand{\eval}[0]{EVAL\xspace}
\newcommand{\finp}[0]{FACT-NP\xspace}

\newcommand{\F}{$\text{F}_1$\xspace}
\newcommand{\K}{\texttt{$\kappa$}\xspace}

\newcommand{\eg}{e.g.,\xspace} 
\newcommand{\ie}{i.e.,\xspace} 

\newcommand\NorecEval{{NoReC$_\text{\textit{eval}}$}\xspace}
\newcommand\NorecFine{{NoReC$_\text{\textit{fine}}$}\xspace}

\title{A Fine-Grained Sentiment Dataset for Norwegian}

\name{Lilja {\O}vrelid, Petter M{\ae}hlum, Jeremy Barnes, Erik Velldal}

\address{University of Oslo\\
  Department of Informatics\\
 {\tt \{liljao,pettemae,jeremycb,erikve\}@ifi.uio.no}\\}

\abstract{
We introduce \NorecFine, a dataset for fine-grained sentiment analysis in Norwegian, annotated with respect to polar expressions, targets and holders of opinion. The underlying texts are taken from a corpus of professionally authored reviews from multiple news-sources and across a wide variety of domains, including literature, games, music, products, movies and more. We here present a detailed description of this annotation effort. We provide an overview of the developed annotation guidelines, illustrated with examples, and present an analysis of inter-annotator agreement. We also report the first experimental results on the dataset, intended as a preliminary benchmark for further experiments.  
\\ \newline  }

\begin{document}

\maketitleabstract

\section{Introduction}

In this work, we describe the annotation of a fine-grained sentiment dataset for Norwegian, analysing opinions in terms of their polar expressions, targets, and holders. The dataset, including the annotation guidelines, is made publicly  available\footnote{\url{https://github.com/ltgoslo/norec_fine}} and is the first of its kind for Norwegian. 
The underlying texts are taken from the Norwegian Review Corpus (NoReC) \cite{Vel:Ovr:Ber:18} -- a corpus of professionally authored reviews across a wide variety of domains, including literature, video games, music, various product categories, movies, TV-series, restaurants, etc. 
In \newcite{Mae:Bar:Ovr:2019}, a subset of the documents, dubbed \NorecEval, were  annotated at the sentence-level, indicating whether a sentence contains an \textit{evaluation} or not. 
These prior annotations strictly indicated evaluativeness and did not include negative or positive polarity, as this can be mixed at the sentence-level. 

In the current work, the previous annotation effort has been considerably extended to include the span of \text{polar expressions} and the corresponding \textit{targets} and \textit{holders} of the opinion. We also indicate the \textit{intensity} of the \textit{positive or negative polarity} on a three-point scale, along with a number of other attributes of the expressions. The resulting dataset, dubbed \NorecFine, comprises almost 8000 sentences (of which roughly half are evaluative) across roughly 300 reviews and includes both subjective and fact-implied evaluations. In addition to discussing annotation principles and examples, we also present the first experimental results. 

The paper is structured as follows. Section~\ref{sec:related} reviews related work, both in terms of related resources for other languages and work on computational modeling of fine-grained opinions. We then go on to discuss our annotation effort in Section~\ref{sec:annotations}, where we describe annotation principles, discuss a number of examples and finally present statistics on inter-annotator agreement. Section~\ref{sec:experiments} presents our first experiments using this dataset for neural machine learning of fine-grained opinions, before Section~\ref{sec:future} discusses some future directions of research. Finally, Section~\ref{sec:summary} summarizes the main contributions of the paper.

\section{Related Work}
\label{sec:related}

Fine-grained approaches to sentiment analysis include opinion annotations as in \cite{Wiebe2005}, aspect-based sentiment \cite{HuandLiu2004}, and targeted sentiment \cite{Vo:2015:TTS:2832415.2832437}. Whereas document- and sentence-level sentiment analysis make the simplifying assumption that all polarity in the text is expressed towards a single entity, fine-grained approaches attempt to model the fact that polarity is directed towards entities (either implicitly or explicitly mentioned). In this section we provide a brief overview of related work, first in terms of datasets and then modeling.

\subsection{Datasets}

One of the earliest datasets for fine-grained opinion mining is the MPQA corpus \cite{Wiebe2005}, which contains annotations of private states in English-language texts taken from the news domain. The authors propose a detailed annotation scheme in which annotators identify subjective expressions, as well as their targets and holders.

Working with sentiment in English consumer reviews, \newcite{Top:Jak:Gur:10} annotate targets, holders and polar expressions, in addition to modifiers like negation, intensifiers and diminishers. The intensity of the polarity is marked on a three-point scale (weak, average, strong). 
In addition to annotating explicit expressions of subjective opinions, \newcite{Top:Jak:Gur:10} annotate \emph{polar facts} that may \emph{imply} an evaluative opinion. A similar annotation scheme is followed by \newcite{Van:Des:Hos:15}, working on financial news texts in Dutch and English, also taking account of implicit expressions of sentiment in polar facts.  

The SemEval 2014 shared task \cite{pontiki-etal-2014-semeval} proposes a different annotation scheme. Given an English tweet, the annotators identify targets, the aspect category they belong to, and the polarity expressed towards the target. They do not annotate holders or polar expressions.

While most fine-grained sentiment datasets are in English, there are datasets available in several languages, such as German \cite{klinger-cimiano-2014-usage}, Czech \cite{steinberger-etal-2014-aspect}, Arabic, Chinese, Dutch, French, Russian, Spanish, Turkish \cite{pontiki-etal-2016-semeval}, Hungarian \cite{szabo2016}, and Hindi \cite{Akhtar2016AspectBS}. Additionally, there has been an increased effort to create fine-grained resources for low-resource languages, such as Basque and Catalan \cite{barnes2018-multibooked}. No datasets for fine-grained SA have previously been created for Norwegian, however. 

\subsection{Modeling}

Fine-grained sentiment is most often approached as a sequence labeling problem \cite{yang-cardie-2013-joint,Vo:2015:TTS:2832415.2832437} or simplified to a classification problem when the target or aspect is given \cite{pontiki-etal-2014-semeval}.

Although the specific architectures vary greatly, state-of-the-art methods for fine-grained sentiment analysis tend to be transfer-learning approaches \cite{chen-qian-2019-transfer}, often using pre-trained language models \cite{peters-etal-2018-deep,devlin-etal-2019-bert} to improve model performance \cite{hu-etal-2019-open}. Additionally, several approaches use multi-task learning to incorporate useful information from similar tasks into fine-grained sentiment models \cite{marasovic-frank-2018-srl4orl,he-etal-2019-interactive,li-etal-2019-unified}. 

Alternatively, researchers have proposed attention-based methods which are adapted for fine-grained sentiment \cite{tang-etal-2019-progressive,bao-etal-2019-attention}. These methods make use of an attention mechanism \cite{bahdanau+al-2014-nmt} which allows the model to learn a weighted representation of sentences with respect to sentiment targets.

Finally, there are approaches which create task-specific models for fine-grained sentiment.  \newcite{liang-etal-2019-aspectguided} propose an aspect-specific gate to improve GRUs.

\section{Annotations}
\label{sec:annotations}

In the following we present our fine-grained sentiment annotation effort in more detail. We provide an overview of the annotation guidelines and present statistics on inter-annotator agreement. The complete set of guidelines is distributed with the corpus.
\begin{figure}[]
\centering
\includegraphics[width=.45\textwidth]{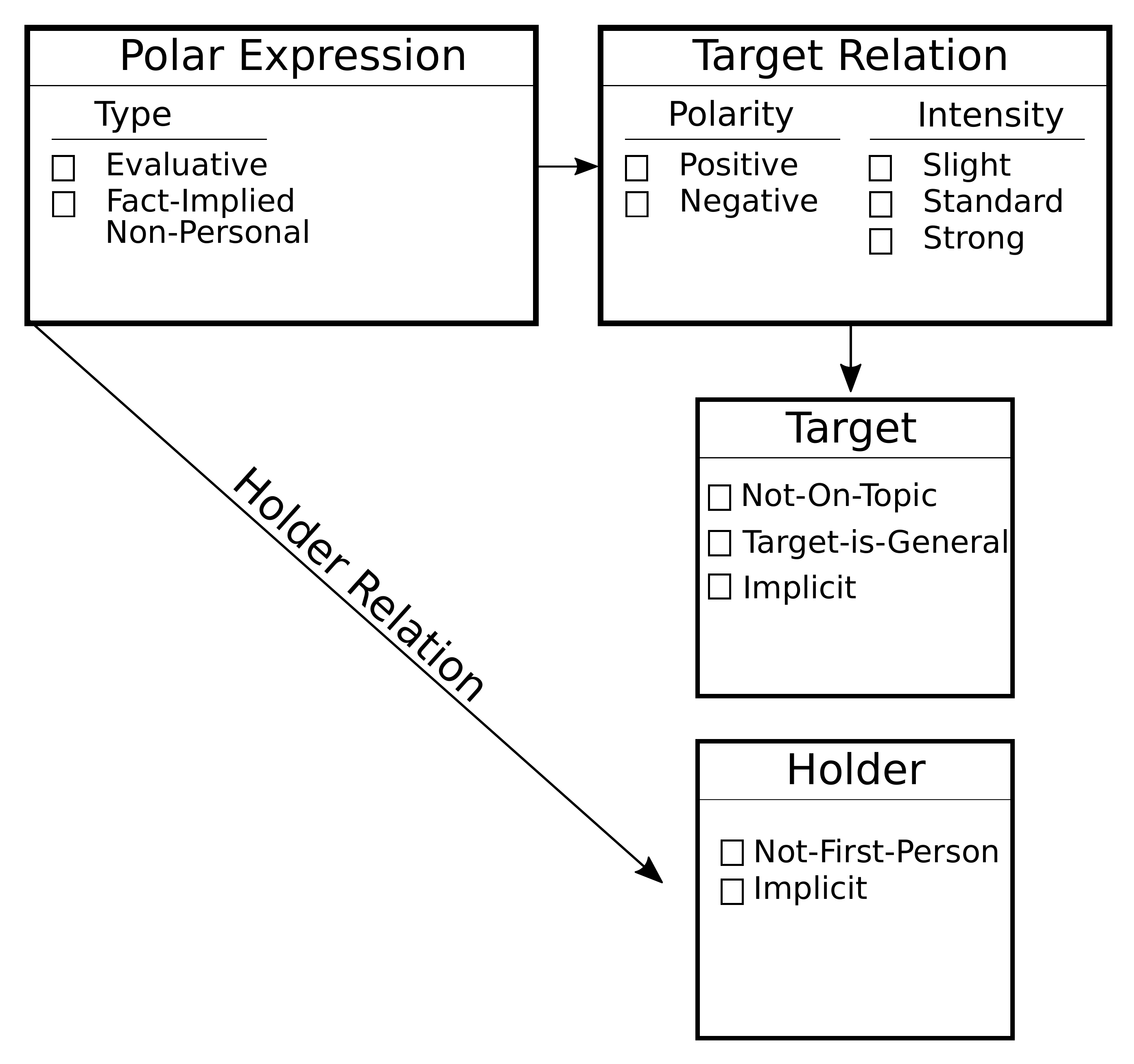}
\caption{Annotation schema for the \NorecFine dataset.}
\label{fig:schema}
\end{figure}

\paragraph{Sentence-level annotations}
We build on the sentence-level annotation of evaluative sentences in the \NorecEval{-corpus} \cite{Mae:Bar:Ovr:2019}, where two types of evaluative sentences were annotated: simple \textit{evaluative} sentences (labeled \eval), or the special case of \textit{evaluative fact-implied non-personal} (\finp) sentences.
The \eval\ label roughly comprises the three opinion categories described by \newcite{Liu:15} as emotional, rational and fact-implied personal. Sentences including emotional responses (arousal) are very often evaluative and involve emotion terms, \eg  \textit{elske} `love', \textit{like} `like', \textit{hate} `hate'. Sentences that lack the arousal we find in emotional sentences may also be evaluative, for instance by indicating worth and utilitarian value, \eg \textit{nyttig} `useful', \textit{verdt (penger, tid)} `worth (money, time)'.
In \NorecEval, a sentence is labeled as \finp\ when it is a fact or a descriptive sentence but evaluation is implied, and the sentence does not involve any personal experiences or judgments.

While previous work \cite{Top:Jak:Gur:10} only annotate sentences that are found to be `topic relevant', \newcite{Mae:Bar:Ovr:2019} choose to annotate all sentiment-bearing sentences, but explicitly include a Not-on-Topic marker. This will allow for assessing the ability of models to reliably identify sentences that are not relevant but still evaluative. 

\begin{figure}
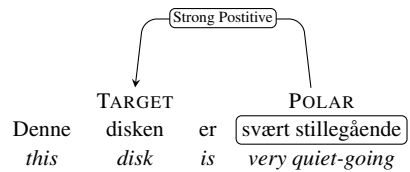

\centering 
\begin{dependency}
\smaller
\begin{deptext}[column sep=.2cm, row sep=.1ex]
\& {\sc Target} \&  \& {\sc Polar} \\
Denne \& disken \& er \& svært stillegående\\
{\it this} \& {\it disk} \& {\it is} \& {\it very quiet-going}\\
\end{deptext}
\wordgroup{2}{4}{4}{polar}

\depedge{4}{2}{Strong Postitive}

\end{dependency}
\caption{Annotation of an \eval sentence (transl. \eng{This disk runs very quietly}).}
\label{fig:eval}
\end{figure}

\begin{figure}
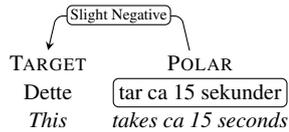

\centering 
\begin{dependency}
\smaller
\begin{deptext}[column sep=.2cm, row sep=.1ex]
  {\sc Target}\& {\sc Polar} \\
Dette \& tar ca 15 sekunder \\
{\it This} \& {\it takes ca 15 seconds}\\
\end{deptext}
\wordgroup{2}{2}{2}{polar}
\depedge{2}{1}{Slight Negative}

\end{dependency}
\caption{Annotation of a \finp sentence (transl. \eng{This takes around 15 seconds}).}
\label{fig:finp}
\end{figure}

\paragraph{Expression-level annotations} 
In our current fine-grained annotation effort we annotate both the \eval and \finp sentences from the \NorecEval corpus. Figure~\ref{fig:schema} provides an overview of the annotation scheme and the entities, relations and attributes annotated. 
Example annotations are provided in Figure~\ref{fig:eval}, for an \eval sentence, and Figure~\ref{fig:finp} for a \finp.
As we can see, positive or negative polarity is expressed by a relation between a \emph{polar expression} and the \emph{target(s)} of this expression and is further specified for its strength on a three-point scale, resulting in six polarity values, ranging from \emph{strong positive} to \emph{strong negative}. The \emph{holder} of the opinion is also annotated if it is explicitly mentioned. Some of the annotated entities are further annotated with attributes indicating, for instance, if the opinion is not \emph{on topic} (in accordance with the topic of the review) or whether the target or holder is \emph{implicit}.

\subsection{Polar Expressions}
A polar expression is the text span that contributes to the evaluative and polar nature of the sentence. For some sentences this may simply be expressed by a sentiment lexeme such as \norex{elsker} \eng{loves}, \norex{forferdelig} \eng{awful} for \eval type expressions. In the case of \finp  polar expressions, any objective description that is seen to reflect the holder's evaluation is chosen, as in Figure~\ref{fig:finp}. Polar  expressions may also include modifiers, including intensifiers such as \norex{very} or modal elements such as \norex{should}. Polar expressions are often adjectives, but verbs and nouns also frequently occur as polar expressions. In our annotation, the span of a polar expression should be large enough to capture all necessary information, without including irrelevant information. In order to judge what is relevant, annotators were asked to consider whether the strength and polarity of the expression 
would change if the span were reduced.

\paragraph{Polar expression span} 
The annotation guidelines further describe a number of distinctions that should aid the annotator in determining the polar expression and its span.
Certain punctuation marks, such as exclamation and question marks, can be used to modify the evaluative force of an expression, and are therefore included in the polar expression if this is the case. 

\begin{figure*}
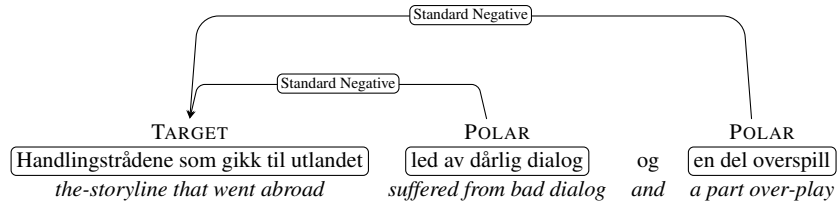

\centering 
\begin{dependency}
\smaller
\begin{deptext}[column sep=.2cm, row sep=.1ex]
{\sc Target} \&  {\sc Polar} \& \& {\sc Polar} \\
Handlingstrådene som gikk til utlandet \& led av dårlig dialog \& og \& en del overspill\\
{\it the-storyline that went abroad} \& {\it suffered from bad dialog} \& {\it and} \& {\it a part over-play}\\
\end{deptext}
\wordgroup{2}{1}{1}{target}
\wordgroup{2}{2}{2}{polar1}
\wordgroup{2}{4}{4}{polar2}

\depedge{2}{1}{Standard Negative}
\depedge{4}{1}{Standard Negative}

\end{dependency}
\caption{An annotated example of a coordinated polar expression (transl. \eng{The storyline that stretches abroad suffered from bad dialog and exaggerated acting}).}
\label{fig:annotex2}
\end{figure*}

Verbs are only included if they contribute to the semantics of the polar expression. For example, in the sentence in Figure~\ref{fig:annotex2} the verb \norex{led} \eng{suffers} clearly contributes to the negative sentiment and is subsequently included in the span of the polar expression.
High-frequent verbs like \norex{å være} \eng{to be} and \norex{å ha} \eng{to have} are generally not included in the polar expression, as shown in the example in Figure~\ref{fig:eval} above.

Prepositions belonging to particle verbs and reflexive pronouns that occur with reflexive verbs are further included in the span. Verbs that signal the evaluation of the author but no polarity are not annotated,
These verbs include \norex{synes} \eng{think} and \norex{mene} \eng{mean}.

Sentence-level adverbials such as \norex{heldigvis} \eng{fortunately}, \norex{dessverre} \eng{unfortunately}, often add evaluation and/or polarity to otherwise non-evaluative sentences. In our scheme, they are therefore annotated as part of the polar expression.

\begin{figure}
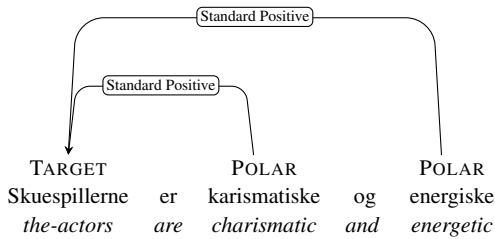

\centering 
\begin{dependency}
\smaller
\begin{deptext}[column sep=.2cm, row sep=.1ex]
{\sc Target}\& \& {\sc Polar} \& \& {\sc Polar} \&\\
Skuespillerne \& er \&   karismatiske \& og \& energiske \& [...] \\
{\it the-actors} \& {\it are} \& {\it charismatic} \& {\it and } \& {\it energetic }\\
\end{deptext}
\depedge{3}{1}{Standard Positive}
\depedge{5}{1}{Standard Positive}

\end{dependency}
\caption{An example of coordinated polar expressions (transl. \eng{The actors are charismatic and energetic}).}
\label{fig:coordinated}
\end{figure}

Coordinated polar expressions are as a general rule  treated as two separate expressions, as in the example in Figure~\ref{fig:coordinated} where there are two conjoined polar expressions with separate target relations to the target. In order to avoid multiple (unnecessary) discontinuous spans, conjunct expressions that share an element, are, however, included in the closest conjunct. An example of this is found in Figure~\ref{fig:annotex2}, where the verbal construction \norex{led av} \eng{suffered from} has both syntactic and semantic scope over both the conjuncts (\norex{led av dårlig dialog} \eng{suffered from bad dialog} and \norex{led av en del overspill} \eng{suffered from some over-play}). 
If the coordinated expression is a fixed expression involving a coordination, the whole expression should be marked as one coherent  entity.

Expletive subjects are generally not included in the span of polar expressions.
Furthermore, subjunctions should not be included unless excluding them alone leads to a discontinous span.

\paragraph{Polar expression intensity}
The intensity of a polar expression is indicated linguistically in several different ways. Some expressions are inherently strongly positive or negative, such as \norex{fabelaktig} \eng{fabulous}, and \norex{katastrofal} \eng{catastrophic}. In other cases, various modifying elements shift the intensity towards either point of the scale, such as adverbs, \eg \norex{uhyre} \eng{immensely} as in \norex{uhyre tynt} \eng{immensely thin}. Some examples of adverbs found with slightly positive or negative expressions are \norex{noe} \eng{somewhat}, \norex{kanskje} \eng{maybe} and \norex{nok} \eng{probably}. The target of the expression can also influence the intensity, and the annotators were urged to consider the polar expressions in context.

\subsection{Targets}
We annotate the targets of polarity by explicitly marking target entities in the text and relating them to the corresponding polar expression via a target relation. In Figure~\ref{fig:eval} for instance we see that the polar expression \norex{svært stillegående} \eng{very quiet-going} is directed at the target \norex{disken} \eng{disk} and expresses a Strong Positive polarity. As a rule of thumb, the span of a target entity should be as short as possible whilst preserving all relevant information. This means that information that does not aid in identifying the target should not be included. When it comes to more formal properties of targets, they are typically nominal, but in theory they can also be expressed through adjectives or verbs. 
Target identification is not always straightforward. Our guidelines therefore describe several guiding principles, as well as some more detailed rules of annotation. For instance, reviewed objects might have easily identifiable physical targets, \eg a tablet can have the targets \textit{screen} and \textit{memory}. However, targets may also have more abstract properties, such as \textit{price} or \textit{ease of use}. A target can also be a property or \emph{aspect} of another target. Following the tablet example above, the target \textit{screen} can have the sub-aspects \textit{resolution}, \textit{color quality}, etc. We can imagine an aspect tree, spanning both upwards and downwards from the object being reviewed.

\paragraph{Canonical targets} Targets are only selected if they are considered \textit{canonical}, meaning that they represent some commonly encountered feature of the object under review. For example, in the sentence \norex{det var en god nerve i hovedmysteriet} \eng{there was a good nerve in the main mystery}, the word \norex{nerve} \eng{nerve} is not annotated as target, as it is not considered to be an essential part of (in this case) all TV series in general. Rather, \norex{god nerve} \eng{good nerve} is annotated as a polar expression targeted at the entity \norex{hovedmysteriet} \eng{the main mystery}.
Fixed expressions, such as \textit{god idé} are a common source of non-canonical expressions. In contrast, in phrases such as \textit{tydelig skuespill} 'clear acting', the target \textit{skuespill} `acting' is seen as an integral part of TV series, and therefore considered so-called canonical.

\paragraph{General targets}
When the polar expression concerns the object being reviewed, we add the attribute \textit{target-is-general}. This applies both when the target is explicitly mentioned in the text and when it is implicit.

The \textit{target-is-general} attribute is not used when a polar expression has a target that is at a lower ontological level than the object being reviewed, as for instance, in the case of the tablet's screen, given our previous example.

\paragraph{Implicit targets}
A polar expression does not need to have an explicit target. Implicit targets are targets that do not appear in the same sentence as the polar expression it relates to. 
We identify three types of implicit targets in our scheme: (i) implicit not-on-topic targets, (ii) implicit general targets and, (iii) implicit canonical aspect targets. A polar expression that refers to something other than what is being reviewed, is marked as \textit{not-on-topic}, even if the reference is implicit. For marking a polar expression that is about the object being reviewed in general, the \textit{target-is-general} attribute is used.

\paragraph{Polar--target combinations}
There are several constructions where targets and polar expressions coincide. Like most Germanic languages, nominal compounding is highly productive in Norwegian and compounds are mostly written as one token. Adjective--noun compounds are fairly frequent and these may sometimes express both polar expression and target in one and the same token, \eg  \norex{favorittfilm} \eng{favourite-movie}. Since our annotation does not operate over sub-word tokens, these types of  examples are marked as polar expressions.

\subsection{Holders}
\begin{figure}
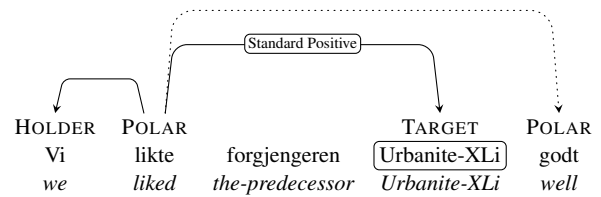

 \centering 
 \begin{dependency}
 \smaller
 \begin{deptext}[column sep=.2cm, row sep=.1ex]
 {\sc Holder} \&  {\sc Polar} \& \& {\sc Target} \& {\sc Polar} \\
Vi \& likte \& forgjengeren \& Urbanite-XLi \& godt \\
{\it we} \& {\it liked} \& {\it the-predecessor} \& {\it Urbanite-XLi} \& {\it well}\\
 \end{deptext}
 \wordgroup{2}{4}{4}{target}
 \depedge[hide label]{2}{1}{}
 \depedge[style=dotted,hide label]{2}{5}{}
 \depedge{2}{4}{Standard Positive}
\end{dependency}
\caption{An annotated example sentence with an explicit holder (transl.\eng{We liked the predecessor, Urbanite-XLi a lot}).}
\label{fig:holder}
\end{figure}

Holders of sentiment are not frequently expressed explicitly in our data, partly due to the genre of reviews, where the opinions expressed are generally assumed to be those of the author. When they do occur though, holders are commonly expressed as pronouns, but they can also be expressed as nouns such as \norex{forfatteren} \eng{the author}, proper names, etc. 
Figure~\ref{fig:holder} shows an annotated example where the holder of the opinion \norex{Vi} \eng{We} is related to a polar expression. Note that this example also illustrates the treatment of discontinuous polar expressions. Discontinuous entities are indicated using a dotted line, as in Figure~\ref{fig:holder} where the polar words \norex{likte} \eng{liked} and \norex{godt} \eng{well} form a discontinuous polar expression.
At times, authors may bring up the opinions of others when reviewing, and in these  cases the holder will be marked with the attribute  Not-First-Person.

\subsection{General}
\begin{figure*}
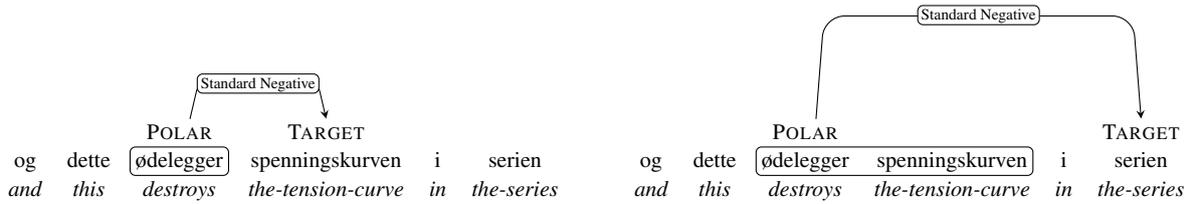

\centering 
\begin{dependency}
\smaller
\begin{deptext}[column sep=.2cm, row sep=.1ex]
\& \& {\sc Polar} \& {\sc Target} \& \&  \\
og \& dette \& ødelegger \& spenningskurven \& i \& serien \\
{\it and} \& {\it this} \& {\it destroys} \& {\it the-tension-curve} \& {\it in} \& {\it the-series} \\
\end{deptext}
\wordgroup{2}{3}{3}{polarI}
\depedge{3}{4}{Standard Negative}
\end{dependency}
\hspace{0.5cm}
\begin{dependency}
\smaller
\begin{deptext}[column sep=.2cm, row sep=.1ex]

\& \&{\sc Polar} \&  \& \& {\sc Target} \\
og \& dette \& ødelegger \& spenningskurven \& i \& serien \\
{\it and} \& {\it this} \& {\it destroys} \& {\it the-tension-curve} \& {\it in} \& {\it the-series} \\
\end{deptext}
\wordgroup{2}{3}{4}{polar}
\depedge{3}{6}{Standard Negative}
\end{dependency}
\caption{Example of a sentence with nested annotation, each level of nesting is shown as a separate annotation layer (transl. \eng{and this destroys the plot development of the series}).}
\label{fig:nesting}
\end{figure*}

We will here discuss some general issues that are relevant for several of the annotated entities and relations.
\paragraph{Nesting}
In some cases, a polar expression and a target together form a polar expression directed at another target. If all targets in these cases are canonical, then the expressions are nested. Figure~\ref{fig:nesting} shows an example sentence where the verb \norex{ødelegger} \eng{destroys} expresses a negative polarity towards the target \norex{spenningskurven} \eng{the tension curve} and the combination \norex{ødelegger spenningskurven} \eng{destroys the tension curve} serves as a polar expression which predicates a negative polarity of the target \norex{serien} \eng{the series}.

\begin{figure*}
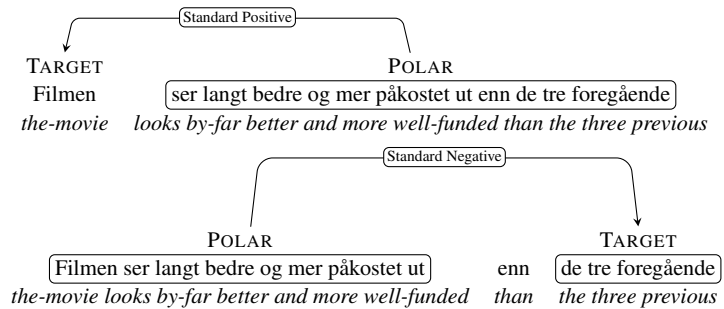

\centering 
\begin{dependency}
\smaller
\begin{deptext}[column sep=.2cm, row sep=.1ex]
{\sc Target}\& {\sc Polar}\\
Filmen \& ser langt bedre og mer påkostet ut enn de tre
foregående\\
{\it the-movie} \& {\it looks by-far better and more well-funded than the three previous }\\
\end{deptext}
\wordgroup{2}{2}{2}{polarI}
\depedge{2}{1}{Standard Positive}

\end{dependency}
\begin{dependency}
\smaller
\begin{deptext}[column sep=.2cm, row sep=.1ex]
{\sc Polar}\& \& {\sc Target}\\
Filmen  ser langt bedre og mer påkostet ut \& enn \& de tre
foregående\\
{\it the-movie looks by-far better and more well-funded} \&  {\it than} \& {\it the three previous }\\
\end{deptext}
\wordgroup{2}{1}{1}{polar}
\wordgroup{2}{3}{3}{target}
\depedge{1}{3}{Standard Negative}

\end{dependency}
\caption{An example of a comparative sentence with nested annotations, here shown as two separate layers of annotation (transl. \eng{The movie looks far better and more expensive than the three previous ones}).}
\label{fig:comparative}
\end{figure*}

\paragraph{Comparatives}
Comparative sentences can pose certain challenges because they involve the same polar expression having relations to two different  targets, usually (but not necessarily) with opposite polarities. Comparative sentences are indicated by the use of comparative adjectival forms, and commonly also by the use of the comparative subjunction  \norex{enn} \eng{than}. In comparative sentences like \norex{X er bedre enn Y} \eng{X is better than Y}, X and Y are entities, and \norex{bedre} \eng{better} is the polar expression. In general we annotate \norex{X er bedre} \eng{X is better} as a polar expression modifying Y, and \norex{bedre enn Y} \eng{better than Y} as a polar expression modifying X. Here there should be a difference in polarity as well, indicating that X is better than Y. The annotated examples in Figure~\ref{fig:comparative} shows the two layers of annotation invoked by a comparative sentence.

\paragraph{Determiners} Demonstratives and articles are generally not included in the span of any expressions, as exemplified by the demonstrative \norex{Denne} \eng{this} in the example in Figure~\ref{fig:eval} above, unless they are needed to resolve ambiguity. Quantifiers such as \norex{noen} \eng{some} , \norex{mange} \eng{many} on the other hand are always included if they contribute to the polarity or intensity of the sentence.

\subsection{Annotation Procedure}

The annotation was performed by several hired student assistants with a background in linguistics and with Norwegian as their native language. All 298 documents in the dataset, comprising 7961 sentences, were annotated independently by two annotators in parallel. 
The doubly annotated documents were then adjudicated by a third annotator different from the two initial annotators. In the initial annotation phase, all annotators were given the possibility to discuss difficult choices in joint annotator meetings, but were encouraged to take independent decisions based on the guidelines if possible. Annotation was performed using the web-based annotation tool Brat \cite{Ste:Pyy:Top:2012}. 

\subsection{Inter-Annotator Agreement}
\label{sec:iaa} 
In this section, we examine inter-annotator agreement. As extracting opinion holders, targets, and opinion expressions at token-level is a difficult task, even for humans \cite{Wiebe2005}, we use soft evaluation metrics, specifically \emph{Binary Overlap} and \emph{Proportional Overlap} \cite{katiyar-cardie-2016-investigating}. Binary Overlap counts any overlapping predicted and gold span as correct. Proportional Overlap instead assigns precision as the ratio of overlap with the predicted span and recall as the ratio of overlap with the gold span, which reduces to token-level \F. Proportional Overlap is therefore a stricter metric than Binary Overlap. Finally, we also compute the Cohen's kappa coefficient \K for Proportional Overlap. We do not do the same for Binary Overlap, as the procedure leads to symmetrically imbalanced predictions (both annotator 1 and annotator 2 are highly likely to have overlapping annotations) and this is known to be problematic for \K \cite{Flight2015-kappa,Zecetal2017-kappaparadox}. In order to calculate inter-annotator agreement for polarity and intensity labels, we report Binary Overlap and Cohen's \K on sentences which have a single polar expression. The reason for this restriction is to avoid having large spans from one annotator overlapping with several smaller spans from a second resulting in either artificially inflated or decreased agreement scores for polarity and intensity.

The inter annotator agreement scores obtained in the first rounds of (double) annotation are reported in Table \ref{tab:inter_annotator}. We find that even though annotators tend to agree on certain parts of the expressions, they agree less when it comes to exact spans.  The proportional and binary holder scores are the highest  (99\% and 92\%  \F). As mentioned earlier, holder expressions tend to be short, often pronominal, hence they are easier to agree on. Targets are more difficult, and may include longer expressions. Further, and as noted during annotation, there is strong agreement on the most central elements of the polar expression, even though there are certain disagreements when it comes to the exact span of a polar expression. 
When it comes to targets, however, there is considerably less agreement among the annotators (73\% Binary \F). More analysis is required in order to fully understand the reasons for this, but it is possible that the distinction between canonical and non-canonical targets has proven difficult to implement in practice. One additional source of disagreement might be the fact that the same target can be mentioned several times in the same sentence, and this might lead to conflicting choices between annotators. 

\begin{table}[t]
    \centering
    \newcommand{\rt}[1]{\rotatebox{90}{#1}}
    \begin{tabular}{@{}llcc@{}}
    \toprule
           & Label &  \F & \K \\
    \cmidrule(l){2-4}
    \multirow{3}{*}{\rt{Prop.}}&Holder &  99\% & 0.64 \\
    &Target &  91\% & 0.57 \\
    &P. Exp.&  77\% & 0.50 \\
        \cmidrule(l){2-4}
    \multirow{3}{*}{\rt{Binary}}&Holder & 92\% & n/a \\
    &Target &  73\% & n/a \\
    &P. Exp.&  90\% & n/a \\
   \bottomrule
    \end{tabular}
    \caption{Inter-annotator agreement \F -scores for holders, targets, and polar expressions. We report \F -scores for  \emph{Proportional Overlap} (percentage of token-level overlap between annotations) and \emph{Binary Overlap} (any overlap between annotations counts towards true positives), as well as Cohen's \K for the proportional scores.}
    \label{tab:inter_annotator}
\end{table}

\begin{table}[t]
    \centering
    \begin{tabular}{@{}ccc@{}}
    \toprule
         & \F & \K \\
         \cmidrule(l){2-3}
         Polarity & 91\% & 0.82 \\
         Intensity & 67\% & 0.28 \\
        \bottomrule
    \end{tabular}
    \caption{Inter-annotator agreement \F and \K scores for polarity and intensity given binary Polar expression overlap. }
    \label{tab:inter_annotator_int_pol}
\end{table}

Table \ref{tab:inter_annotator_int_pol} presents inter-annotator agreement for polarity (positive or negative) and intensity (strong, standard and slight) separately. We find that whereas the general polarity shows fairly high agreement (91\% \F and 0.82 \K), the annotation of intensity invokes considerably less agreement (67\% \F and 0.28 \K) among the annotators.

\section{Corpus Statistics}
\label{sec:corpus}

Table \ref{tab:stats} presents some relevant statistics for the resulting \NorecFine dataset, providing the distribution of sentences, as well as holders, targets and polar expressions in the train, development and test portions of the dataset, as well as the total counts for the dataset as a whole. We also report the average length of the different annotated categories.
As we can see, the total of 7961 sentences that are annotated comprise 7581 polar expressions, 5999 targets, and 735 holders. In the following we present and discuss some additional core statistics of the annotations. 

\begin{figure}
    \centering
    \includegraphics[width=0.7\linewidth]{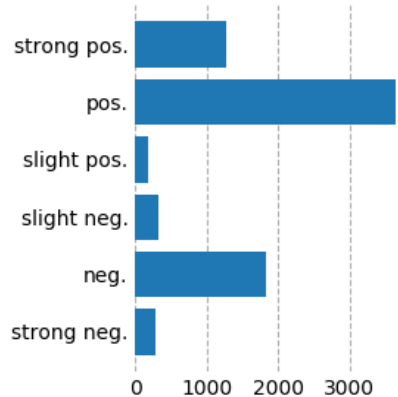}
    \caption{Distribution of labels and intensities.}
    \label{fig:polarity_dist}
\end{figure}

\paragraph{Distribution of valence, intensity and polar expressions}
Figure~\ref{fig:polarity_dist} plots the distribution of polarity labels and their intensity scores. We see that the intensities are clearly dominated by \textit{standard} strength, while there are also 1270 \textit{strong} labels for positive. Regardless of intensity, we see that positive valence is more prominent than negative, and this reflects a similar skew for the document-level ratings in this data  \cite{Vel:Ovr:Ber:18}.  

The \textit{slight} intensity is infrequent, with 190 positive and 330 negative polar expressions with this label. The relative difference here can be explained by the tendency to hedge negative statements more than positive ones \cite{kiritchenko-mohammad-2016-effect}. \textit{Slight negative} is the minority class, with only 190 examples, followed by \textit{strong negative} with 292 examples. Overall, the distribution of intensity scores in \NorecFine is very similar to what is reported for other fine-grained sentiment datasets for English and Dutch \cite{Van:Des:Hos:15}.

As we can see from Table~\ref{tab:stats}, the average number of tokens spanned by a \textit{polar expression} is 4.6. 
Interestingly, if we break this number down further, we find that the negative expressions are on average longer than the positives for all intensities: while the average length of negative expressions are 5.5, 5.4, and 5.7 tokens for   \textit{standard}, \textit{strong}, and \textit{slight} respectively, the corresponding counts for the positives are 4.2, 4.3, and 4.8. 
Overall, we see that the \textit{slight} examples are the longest, often due to hedging strategies which include adverbial modifiers, \eg `a bit' or `maybe'.

Finally, note that only 380 of the annotated polar expressions are of the type \textit{fact-implied non-personal}. 

\paragraph{Distribution of holders and targets}
Returning to the token counts in  Table~\ref{tab:stats}, we see that while references to \textit{holders} are just one word on average (often just a pronoun), \textit{targets} are two on average. 
However, not all targets and holders have a surface realization. 
There are 6846 polar expressions with an \textit{implicit holder}, 1426 with an \textit{implicit target}, and 2186 with the tag \textit{target-is-general}. 

Finally, we note that there are 925 examples where the target is further marked as \textit{not-on-topic} and 150 where the holder is \textit{not-first-person}.  

\begin{table}[]
    \centering
    \begin{tabular}{@{}lrrrrr@{}}
    \toprule 
    & \multicolumn{4}{c}{\# Examples}  \\
    \cmidrule(lr){2-5}
    & Train & Dev. & Test & Total & Avg. len.\\
    \cmidrule(l){2-6}
    Sents.  & 5915 & 1151 &895 &7961 & 16.8\\
    Holders & 584& 76 & 75& 735& 1.1\\
    Targets & 4458&832 &709 &5999 & 2.0\\
    Polar exp. &5659 & 1050 & 872 &7581 & 4.6\\
    \bottomrule
    \end{tabular}
    \caption{Number of sentences and annotated holders, targets and polar expressions across the data splits. The final column shows average token length. (Holders and targets can also be \emph{implicit}, but these are not counted in this table.)}
    \label{tab:stats}
\end{table}

\section{Experiments}
\label{sec:experiments}

To provide an idea of the difficulty of the task, here we report some preliminary experimental results for the new dataset, intended as benchmarks for further experiments. Casting the problem as a sequence labeling task, we train a model to jointly predict holders, targets and polar expressions. Below, we first describe the evaluation metrics and the experimental setup, before finally discussing the results.

\subsection{Experimental Setup}

We train a Bidirectional LSTM with a CRF inference layer, which has shown to be competitive for several other sequence labeling tasks \cite{HuangXY15,lample-etal-2016-neural,panchendrarajan-amaresan-2018-bidirectional}. We use the IOB2 label encoding for sources, targets, and polar expressions, including the binary polarity of the latter, giving us nine tags in total.\footnote{The tagset is \{O, B-Holder, I-Holder, B-Target, I-Target, B-Pos, I-Pos, B-Neg, I-Neg\}.} This naturally leads to a lossy representation of the original data, as the relations, nested annotations, and polar intensity are ignored. 

Our model uses a single BiLSTM layer (100 dim.) to extract features and then a CRF layer to make predictions. We train the model 
using Adam \cite{Kingma2014a} for 40 epochs with a patience of 5, and use dropout to regularize both the BiLSTM (0.5) and CRF (0.3) layers. The word embeddings are 100 dimensional fastText SkipGram  \cite{bojanowski2016enriching} vectors 
 trained on the NoWaC
corpus \cite{guevara-2010-nowac} 
and made available from the NLPL vector repository \cite{Far:Kut:Oep:17}.\footnote{Corresponding to model ID 124 from the repository:\\   \url{http://vectors.nlpl.eu/repository/}} The pre-trained embeddings are further fine-tuned during training. We report held-out test results for the model that achieves the best performance on the development set and use the standard train/development/test split provided with the dataset (shown in Table \ref{tab:stats}).
All results are reported using the Proportional and Binary precision, recall and \F scores, computed as described in Section~\ref{sec:iaa} above. 

\subsection{Results}

\begin{table}[]
    \centering
    \newcommand{\rt}[1]{\rotatebox{90}{#1}}
    \begin{tabular}{@{}lllll@{}}
    \toprule
           & & P & R & \F \\
    \cmidrule(l){2-5}
    \multirow{3}{*}{\rt{Prop.}}&Holder & 56.4 & 34.3 & 42.4 \\
    &Target & 28.2 & 36.4 & 31.3 \\
    &P. Exp.& 28.1 & 36.4 & 31.3 \\
    \cmidrule(l){2-5}
    \multirow{3}{*}{\rt{Binary}}&Holder & 55.8 & 35.1 & 43.5 \\
    &Target & 38.4 & 40.4 & 39.1 \\
    &P. Exp.& 66.3 & 57.7 & 61.5 \\
   \bottomrule
    \end{tabular}
    \caption{Precision (P), recall (R), and Micro \F for holders, targets, and polar expressions. Prop. refers to proportional overlap and Binary refers to binary overlap.}
    \label{tab:benchmark_results}
\end{table}{}

Table \ref{tab:benchmark_results} shows the results of the proportional and binary Overlap measures for precision, recall, and \F. The baseline model achieves modest results when compared to datasets that do not involve multiple domains \cite{yang-cardie-2013-joint,barnes2018-multibooked}, with 42.4, 31.3, and 31.3 Proportional \F on holders, targets, and polarity expressions, respectively (43.5, 39.1, 61.5 Binary \F). However, this is still better than previous results on cross-domain datasets \cite{ding2017}. The domain variation between documents leads to a lower overlap between holders, targets, and polar expressions seen in training and those at test time (56\%, 28\%, and 50\%, respectively). We argue, however, that this is a more realistic situation regarding available data, and that it is important to move away from simplifications where training and test data are taken from the same distribution.

\section{Future Work}
\label{sec:future}

In follow-up work we plan to further enrich the annotations with additional compositional information relevant to sentiment, most importantly negation but also other forms of valence shifters. Although our data already contains multiple domains, it is still all within the genre of reviews, and while we plan to test cross-domain effects within the existing data we would also like to add annotations for other different genres and text types, like editorials. 

In terms of modeling, we also aim to investigate approaches that better integrate the various types of annotated information (targets, holders, polar expressions, and more) and the relations between them when making predictions, for example in the form of multi-task learning. Modeling techniques employing attention or aspect-specific gates that have provided state-of-the-art results for English provide an additional avenue for future experimentation.

\section{Summary}
\label{sec:summary}

This paper has introduced a new dataset for fine-grained sentiment analysis,  
the first such dataset available for Norwegian. 
The data, dubbed \NorecFine, comprise a subset of documents in the Norwegian Review Corpus, a collection of professional reviews across multiple domains. 
The annotations mark polar expressions with positive/negative valence together with an intensity score, in addition to the holders and targets of the expressed opinion. Both subjective and objective expressions can be polar, and a special class of objective expressions called \textit{fact-implied non-personal} expressions are given a separate label. The annotations also indicate whether holders are first-person (\ie the author) and whether targets are on-topic. Beyond discussing the principles guiding the annotations and describing the resulting dataset, we have also presented a series of first classification results, providing benchmarks for further experiments. The dataset, including the annotation guidelines, are made publicly available.\footnote{\url{https://github.com/ltgoslo/norec_fine}} 

\section*{Acknowledgements}
This work has been carried out as part of the SANT project (Sentiment Analysis for Norwegian Text), funded by the Research Council of Norway (grant number 270908). We also want to express our gratitude to the annotators: Tita Enstad, Anders Næss Evensen, Helen Ørn Gjerdrum, Petter Mæhlum, Lilja Charlotte Storset, Carina Thanh-Tam Truong, and Alexandra Wittemann.

\section*{Bibliographical References}


\end{document}